\documentclass[conference]{IEEEtran}
\IEEEoverridecommandlockouts
\usepackage{amsthm,amsmath}
\usepackage[utf8]{inputenc}
\usepackage{import}
\usepackage{graphicx}
\usepackage{rotating}
\usepackage[graphicx]{realboxes}
\graphicspath{ {images/} }
\usepackage{array}
\usepackage[citestyle=numeric-comp,bibstyle=ieee,%
backend=biber,sortcites]{biblatex}
\usepackage{orcidlink}
\usepackage{tabularx}
\usepackage{booktabs}
\newcolumntype{Y}{>{\raggedright\arraybackslash}X}

\renewcommand{\arraystretch}{0.85}
\usepackage{enumitem}
\setlist{nosep}
\usepackage{tikz}
\usetikzlibrary{arrows.meta,positioning,fit}
\definecolor{LineCol}{HTML}{000000}
\tikzset{
  box/.style={draw=LineCol, fill=white, rounded corners=2pt, align=center, inner sep=3pt, minimum width=24mm, minimum height=8mm, line width=0.5pt},
  dsbox/.style={draw=LineCol, fill=white, rounded corners=2pt, align=center, inner sep=1.5pt, minimum width=16mm, line width=0.4pt, font=\scriptsize},
  grp/.style={draw=LineCol, rounded corners=2pt, inner sep=3pt, line width=0.4pt},
  ar/.style={-Latex, draw=LineCol, line width=0.5pt}
}
\begin{document}

\title{Human-Grounded Calibration for\\ Long-Text Image--Text Congruence in Vision-Language Models\thanks{This work was funded by Fundação para a Ciência e a Tecnologia (UID/00124/2025, UID/PRR/124/2025, PRT/BD/154926/2023, Nova School of Business and Economics, DOI: https://doi.org/10.54499/UID/00124/2025) and LISBOA2030 (DataLab2030 - LISBOA2030-FEDER-01314200). ChatGPT and Claude were used to polish and edit each section of the paper. }}

\author{
    \IEEEauthorblockN{Alessandro Gambetti\orcidlink{0000-0003-3389-3784}}
    \IEEEauthorblockA{\textit{NOVA School of Science and Technology} \\
    \textit{Universidade NOVA de Lisboa}\\
    Caparica, Portugal \\
    a.gambetti@campus.fct.unl.pt}
    \and
    \IEEEauthorblockN{Qiwei Han\orcidlink{0000-0002-6044-4530}}
    \IEEEauthorblockA{\textit{Nova School of Business and Economics} \\
    \textit{Universidade NOVA de Lisboa}\\
    Carcavelos, Portugal \\
    qiwei.han@novasbe.pt}
}

\maketitle
\vspace{-1cm}
\begin{abstract}
Long-text image--text congruence scoring is increasingly important for vision-language systems that must evaluate whether detailed textual descriptions match visual content. However, raw similarity scores from dual-encoder models are difficult to interpret as calibrated congruence measures, especially under the modality gap between image and text embeddings. This paper proposes Congruency Score (CS), a lightweight calibration layer that maps image--text similarity evidence into a bounded score. Using DOCCI and Urban1k, we evaluate four frozen vision-language backbones and show that observed reductions in post-projection centroid distance do not uniformly improve image--text retrieval performance. Human-grounded evaluations on DOCCI further reveal a trade-off: direct post-hoc calibration preserves high association with human judgments, whereas selected projection-based configurations can reduce threshold-relevant slope and intercept distortions at the cost of retrieval performance and association strength. These results establish long-text image--text congruence scoring as a calibrated score-estimation problem, where retrieval performance, human association, and threshold calibration must be evaluated as distinct objectives. CS provides a lightweight way to expose and operationalize this separation.
\end{abstract}
\begin{IEEEkeywords}
image--text matching, vision-language models, modality gap, score calibration, reference-free evaluation, human-grounded evaluation
\end{IEEEkeywords}

\section{Introduction}
Vision-language systems are commonly evaluated through image--text retrieval, caption generation, or downstream multimodal task performance \cite{radford2021learning,li2022blip,zhai2023sigmoid,hessel2021clipscore}. However, many practical settings require a more specific capability: estimating whether a long textual description is semantically congruent with an image. This problem differs from standard image-sentence retrieval because the text may contain detailed attributes, spatial relations, object counts, contextual descriptions, and fine-grained constraints. Recent benchmarks in dense captioning, phrase grounding, and compositional image--text reasoning show that vision-language models often struggle when evaluation requires grounding attributes, relations, ordering, and partially correct descriptions, not merely retrieving the paired caption \cite{plummer2015flickr30kentities,krishna2017visualgenome,ponttuset2020localized,thrush2022winoground,yuksekgonul2022aro,urbanek2024picture}. In such settings, a model must not only rank candidate pairs, but also produce scores that can be interpreted as calibrated evidence of image--text congruence.
Long-text image--text congruence scoring is increasingly relevant because many real vision-language applications involve textual evidence that is richer than a short caption. In medical and scientific settings, an image may be paired with a report-like description containing locations, findings, uncertainty markers, and negated observations. In e-commerce, a product image may be accompanied by a multi-sentence product description or customer review that mentions materials, size, color, usage conditions, and subjective claims. In news, social media, and creative-generation workflows, visual content may be associated with article excerpts, posts, or generated briefs that contain entities, events, and contextual assertions only partially visible in the image. These settings exceed the short-caption regime for which many contrastive dual encoders were originally trained; for example, the canonical CLIP text encoder supports only 77 tokens, motivating long-context variants such as LongCLIP and JinaCLIP \cite{radford2021learning,zhang2024long,koukounas2024jina}. Even when longer-context models reduce truncation, raw dual-encoder similarity remains a retrieval-oriented score: it may rank a paired item above distractors without providing an interpretable estimate of congruence. We therefore treat long-text image--text evaluation as a calibrated score-estimation problem, where the goal is not only to order candidate pairs correctly, but to produce a bounded score whose scale is meaningful for thresholding, filtering, and cross-instance comparison.

Dual-encoder contrastive models such as CLIP and its variants have advanced image-sentence retrieval by mapping images and text into a shared embedding space \cite{radford2021learning}. Reference-free metrics such as CLIPScore \cite{hessel2021clipscore} reuse this similarity to assess image--text relevance without ground-truth captions. Yet several structural limitations constrain their use as calibrated congruence measures in long-text settings. First, contrastive models exhibit a well-documented \textit{modality gap}, in which image and text embeddings occupy separated regions of the latent space \cite{liang2022mind}, distorting similarity-based scores. Second, widely used CLIP variants impose short context windows (e.g., 77 tokens), forcing truncation when descriptions span multiple sentences or claims \cite{zhang2024long,urbanek2024picture}. Third, cosine similarity is primarily optimized for relative retrieval and need not be meaningful as an absolute semantic score \cite{steck2024cosine}. Fourth, a retrieval score can preserve the correct ordering of candidates while remaining poorly calibrated against human judgments. These issues motivate our focus on calibration and threshold-relevant score behavior, not only on retrieval accuracy.
To address these limitations, we propose the Congruency Score (CS), a lightweight, reference-free calibration layer for long-text image--text congruence. CS introduces trainable projection maps on top of frozen image and text encoders, followed by a monotonic calibration layer that maps similarity logits into a bounded score. The effect of these projections on centroid distance is measured ex post rather than imposed through an explicit gap-reduction loss. CS makes this objective separation measurable: it reveals when retrieval rank, association with human judgments, and threshold calibration agree, and when they diverge.
This paper makes three contributions:
\begin{itemize}
    \item We formulate long-text image--text congruence scoring as a calibrated score-estimation problem. This formulation makes explicit that retrieval metrics such as MRR and Recall@1 do not determine whether a score is numerically meaningful for thresholding or cross-instance comparison.
    \item We empirically show that post-projection centroid-distance changes and retrieval effectiveness are not mechanically coupled. Across four frozen vision-language backbones and two long-text datasets, observed reductions in centroid distance can coincide with lower MRR and Recall@1, demonstrating that representation proximity, retrieval rank, and score calibration capture different properties of a vision-language system.
    \item We evaluate CS as a lightweight calibration layer that makes this separation measurable. Human-grounded evaluation shows that direct post-hoc calibration yields high association with HGT and low pointwise error, while selected projection-based configurations reduce threshold-relevant slope and intercept distortions at the cost of retrieval performance and association strength.
\end{itemize}
\section{Related Work}
\label{sec:related_work}
\subsection{Vision-Language Retrieval and Long-Text Image--Text Matching}
Dual-encoder contrastive models such as CLIP \cite{radford2021learning} and sigmoid-loss variants such as SigLIP \cite{zhai2023sigmoid} have become standard for image-sentence retrieval, and fusion-encoder models such as ALBEF \cite{li2021align} and BLIP \cite{li2022blip} add cross-modal interaction for finer-grained matching. A systematic survey of cross-modal retrieval \cite{wang2025cross} documents that these models are predominantly trained and evaluated on short captions. This short-caption bias propagates a practical constraint: many backbones cap text length (e.g., 77 tokens), and recent work shows that dense or long captions degrade CLIP-style matching \cite{urbanek2024picture}. LongCLIP \cite{zhang2024long} extends the supported text length and JinaCLIP \cite{koukounas2024jina} adopts a long-context text encoder, both targeting longer inputs. These efforts improve the \emph{ranking} of long-text pairs. This need is aligned with recent benchmarks that move beyond short-caption retrieval toward dense descriptions, phrase grounding, and compositional image--text reasoning \cite{plummer2015flickr30kentities,krishna2017visualgenome,ponttuset2020localized,thrush2022winoground,yuksekgonul2022aro}. Our concern is different: even when long-text pairs are ranked correctly, the underlying similarity may be a poor \emph{absolute} congruence signal, which is the calibration problem we study.
\subsection{Reference-Free Image--Text Evaluation}
CLIPScore \cite{hessel2021clipscore} introduced a widely used reference-free image--text metric based on cosine similarity in CLIP's embedding space, removing the dependence on ground-truth captions. Earlier image-caption evaluation metrics largely focused on lexical overlap or reference-based consensus, including BLEU \cite{papineni2002bleu}, CIDEr \cite{vedantam2015cider}, and SPICE \cite{anderson2016spice}. These metrics established the importance of human-consistent evaluation but depend on reference captions or semantic parses, and TIGEr further introduced text-to-image grounding into caption evaluation \cite{jiang2019tiger}. CLIPScore moved evaluation into a reference-free dual-encoder setting, but it still inherits the geometry and calibration properties of the underlying embedding similarity. Subsequent analysis shows that cosine similarity in learned embedding spaces can be sensitive to geometry and normalization, and is not guaranteed to track semantic relatedness \cite{steck2024cosine}. Reference-free metrics therefore inherit whatever miscalibration is present in the underlying embedding similarity. We retain the reference-free setting but treat the raw similarity as an input signal to an explicit calibration layer, rather than as a final congruence score, so that its magnitude can be interpreted on a bounded, thresholdable scale.
%
%
{\color{black} Because dual-encoder cosine similarity optimizes retrieval ranking over absolute semantic scaling, its compressed range fails to track partial long-text mismatches. We therefore treat raw similarity as evidence to be calibrated against human judgments rather than a final metric.}

\subsection{The Modality Gap and Contrastive Gap}
Liang et al.\ \cite{liang2022mind} characterize the modality gap as a structural separation between image and text embeddings, attributable to initialization and contrastive optimization, and quantify it by the distance between modality centroids. Follow-up work refines this account: Shi et al.\ \cite{shi2023towards} analyze why the gap arises in CLIP, Fahim et al.\ \cite{fahim2024modgap} argue that the relevant phenomenon is a broader contrastive gap rather than a purely modality-specific one, and Jiang et al.\ \cite{jiang2023understanding} study latent modality structures in multimodal representation learning. This literature is primarily concerned with representation geometry. We instead ask whether projection-induced changes in centroid distance, measured ex post inside a scoring layer, correspond to better retrieval or better calibrated congruence scores.
\subsection{Score Calibration and Human-Grounded Evaluation}
Mapping a model score to a probability-like value is a calibration problem. Platt scaling \cite{platt1999probabilistic} fits a scalar sigmoid transform to a score; isotonic regression and related methods provide nonparametric alternatives. The calibration literature distinguishes discrimination from numerical reliability: a model can rank examples correctly while producing poorly calibrated scores \cite{zadrozny2002transforming,niculescu2005predicting,guo2017calibration}. Temperature scaling is a restricted post-hoc calibration method, whereas isotonic and binning-based methods provide more flexible alternatives \cite{naeini2015obtaining}. Because our target is a continuous human judgment rather than binary correctness, our use of MAE, MSE, and binned calibration curves is closer to continuous-target calibration than to standard classification ECE \cite{kuleshov2018calibrated}. Calibration is well studied for classifier confidence, but is comparatively underexplored for reference-free multimodal congruence under long text, where the target is a continuous human judgment rather than a binary label. We adopt a Platt-style scalar calibration on a learned image--text similarity and evaluate alignment with human-grounded labels using correlation, mean error, binned calibration curves, and orthogonal regression. Hard negative sampling, which improves contrastive discrimination \cite{robinson2021contrastive,mikolov2013distributed}, is used during training to stress-test the scoring layer on near-miss mismatches.
\noindent
Our work lies at the intersection of these lines of research: it treats long-text image--text congruence as a calibrated score-estimation problem under modality-gap constraints, rather than as a pure retrieval-ranking problem.
\section{Method}
\label{sec:model_architecture}
\begin{figure*}[t]
  \centering
  \includegraphics[width=0.8\linewidth]{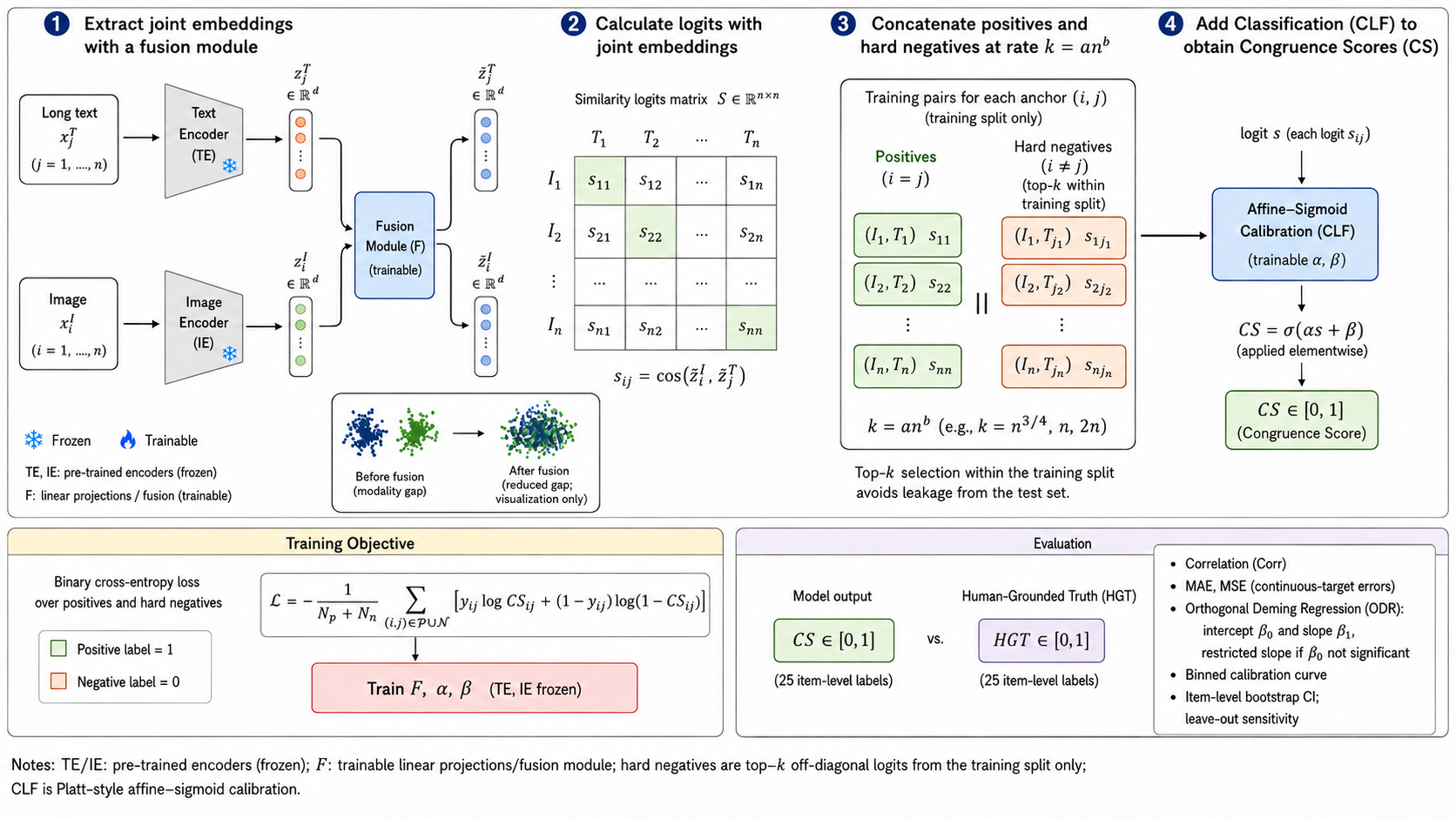}
  \caption{Congruency Score (CS) pipeline: frozen dual encoders produce modality-specific embeddings; trainable linear projections map image and text embeddings into a scoring space; and a Platt-style affine--sigmoid calibration maps similarity logits to a bounded $CS \in [0,1]$. The effect of the projections on centroid distance is measured ex post.}
  \label{fig:architecture}
\end{figure*}
Figure~\ref{fig:architecture} illustrates the proposed CS pipeline. We build on a dual-encoder design with a frozen Text Encoder $f_T$ and a frozen Image Encoder $f_I$, and learn a lightweight scoring layer on top of their outputs.
\paragraph{Step 1: Projection into a Shared Scoring Space}
For an image--text pair $i$, the frozen encoders yield modality-specific embeddings
\begin{equation}
z_i^I = f_I(x_i^I), \qquad z_i^T = f_T(x_i^T),
\end{equation}
which frequently remain separated in the latent space due to the modality gap. {\color{black} A trainable fusion projection module $F$ applies a linear map  $W$} to map each modality into a shared scoring space:
\begin{equation}
\tilde z_i^I = W z_i^I, \qquad \tilde z_i^T = W z_i^T .
\end{equation}
The maps are trained through the paired classification objective (Step 4); we do not impose an explicit centroid-distance loss, and instead measure their effect on the modality gap ex post using centroid distance.
\paragraph{Step 2: Pairwise Similarity Logits}
After $\ell_2$ normalization, we compute cosine similarity between projected embeddings,
\begin{equation}
s_{ij} = \cos\!\big(\tilde z_i^I, \tilde z_j^T\big),
\end{equation}
and treat $s_{ij}$ as the logit for downstream calibration. This scalar is compatible with ranking-based retrieval while remaining amenable to calibration.
\paragraph{Step 3: Hard Negative Construction}
Practical failure cases involve \emph{near-miss} mismatches that are superficially plausible but semantically inconsistent. We construct training pairs by concatenating positive logits $s_{ii}$ with hard negatives $s_{ij}$ ($i \neq j$). Hard negatives are sampled exclusively from within the training split to avoid leakage from the test pool. We sample the top-$k$ most similar negatives, with $k = a n^b$ controlling negative volume relative to the training-set size $n$.
\paragraph{Step 4: Platt-Style Calibration}
We map the similarity logit to a bounded score with a scalar affine--sigmoid transform,
\begin{equation}
CS_{ij} = \sigma\!\big(\alpha\, s_{ij} + \beta\big),
\end{equation}
which is monotonic in $s_{ij}$ and therefore preserves the similarity ranking while producing an interpretable score in $[0,1]$. Here $f_I, f_T$ are frozen; {\color{black} $W$ is a linear projection map } ($d \!\rightarrow\! d$); and $(\alpha,\beta)$ are Platt-style scalar calibration parameters. 
We deliberately keep this parameterization minimal: the projection maps is linear and the calibration is a scalar affine--sigmoid transform, analogous to Platt-style calibration \cite{platt1999probabilistic} applied to a learned image--text similarity signal. 
{\color{black} We chose a linear projection into a shared space not only because it has been shown that modality is a linearly separable phenomenon  \cite{fahim2024modgap}, but also to reduce potential overfitting. 
}
This minimalism keeps CS pluggable and leaves upstream encoders untouched, since the entire mapping from similarity evidence to the calibrated congruence score is described by a small, inspectable set of parameters.

\paragraph{Training Objective}
Let $\mathcal{P}=\{(i,i)\}$ denote matched image--text pairs and let $\mathcal{N}$ denote the top-$k$ off-diagonal hard negatives selected from the training split. For each pair $(i,j)$ we set $y_{ij}=1$ if $(i,j)\in\mathcal{P}$ and $y_{ij}=0$ if $(i,j)\in\mathcal{N}$. The parameters are trained under binary cross-entropy:
\begin{equation}
\mathcal{L}
=
-\!\!\sum_{(i,j)\in \mathcal{P}\cup\mathcal{N}}\!\!
\big[
y_{ij}\log CS_{ij}
+
(1-y_{ij})\log(1-CS_{ij})
\big].
\end{equation}
This objective trains the {\color{black} linear projection $W$ and the affine--sigmoid parameters $(\alpha,\beta)$ to discriminate matched pairs from hard mismatches. Our approach relies on self-supervised learning, which makes it reference-free from external references such as human labeling.
The projection is thus trained through the paired classification objective; its effect on the modality gap is not imposed as an explicit centroid-distance penalty but is measured ex post using centroid distance.}
\paragraph{Calibration Baselines}
Because the proposed calibration layer is intentionally minimal, we situate it against simpler monotonic alternatives. Temperature-only scaling, $CS=\sigma(s/\tau)$ with $\tau>0$ fitted on the training split, rescales the similarity logit but cannot correct an intercept offset. The affine--sigmoid calibration used in ZS-CLF and projection+CLF, $CS=\sigma(\alpha s+\beta)$, adds an intercept term and can therefore correct both under- or over-confidence (through $\alpha$) and systematic baseline offset (through $\beta$). This distinction is central to our analysis, since two models can induce similar rankings while differing in threshold behavior; we accordingly report calibration geometry through slope and intercept estimates alongside continuous-target error, rather than treating rank preservation alone as evidence of score reliability. We do not fit nonparametric or flexible calibration models such as isotonic regression, spline calibration, or beta calibration directly on the HGT labels, because doing so would use the human evaluation set for model fitting and would be severely underdetermined at this sample size. Our reported comparisons focus on transformations and projection-based variants that do not use HGT as supervised training data; extending the HGT set would make held-out human-supervised isotonic, spline, or beta calibration a next step.

\section{Experiments}
\subsection{Experimental Setup}
\label{sec:exp_setup}

\noindent\textbf{Datasets.}
We evaluate CS on two long-text multimodal datasets: DOCCI \cite{onoe2024docci} (15,000 image--text pairs, $\sim$134 words per description) and Urban1k \cite{zhang2024long} (1,000 pairs, $\sim$122 words). Because Urban1k provides no official split, we partition it 80/20 into train/test and treat it as a smaller robustness dataset rather than the primary basis for generalization.

\noindent\textbf{Backbone Models.}
We use four frozen dual-encoder backbones: (i) JinaCLIP ($d{=}768$) \cite{koukounas2024jina} with a BERT-based text encoder (8,192 tokens) and an EVA02-based image encoder; (ii) CLIP-ViT-L/14 ($d{=}768$) \cite{radford2021learning}, limited to 77 tokens; (iii) SigLIP-ViT-B-16-384 ($d{=}1152$) \cite{zhai2023sigmoid}, limited to 64 tokens; and (iv) LongCLIP-ViT-L/14 ($d{=}768$) \cite{zhang2024long}, supporting 248 tokens.

This selection lets us separate two factors often conflated in long-text image--text evaluation: the ability to encode longer text and the ability to produce calibrated congruence scores.

\noindent\textbf{Implementation Details.}
We train the projection maps and the calibration parameters jointly with binary cross-entropy and AdamW (learning rates $10^{-4}$ for projection, $10^{-3}$ for calibration; early stopping, patience 50). After early stopping, we freeze the projection maps, reinitialize the calibration, and refit it with L-BFGS, selecting calibration hyperparameters by validation F1. We evaluate three negative sampling ratios: $k=n^{3/4}$ ($a{=}1,b{=}3/4$) \cite{mikolov2013distributed}, $k=n$ ($a{=}1,b{=}1$), and $k=2n$ ($a{=}2,b{=}1$).

\noindent\textbf{Modality Gap.}
We quantify the modality gap as the squared distance between modality centroids \cite{liang2022mind}: $\|C^{I}-C^{T}\|^2$, with $C^m = \frac{1}{n}\sum_{j}E_j^m$ and values closer to 0 indicating a smaller gap.

\noindent\textbf{Retrieval Metrics.}
We report Mean Reciprocal Rank (MRR) and Recall@1 (R@1) for image-to-text (I2T) retrieval. We focus on I2T because the target use case is long-text congruence scoring for a given image--description pair, where the score is consumed as image-conditioned textual grounding evidence rather than as a symmetric retrieval benchmark.

\noindent\textbf{Human-Grounded Truth (HGT).}
To evaluate alignment with human perception on DOCCI, we sample 10 positive and 10 negative pairs (random mismatched texts), plus 5 positive pairs with text partially perturbed via \textit{GPT-4o-mini} to broaden score coverage. We add two attention checks and randomize all 27 items. Participants were instructed to judge whether the textual description was visually supported by the image, including objects, attributes, spatial relations, counts, and scene context, rather than whether the text was fluent or plausible in isolation. Participants rate congruence on a 0--100 ordinal scale (0/20/40/60/80/100) with a ``Cannot decide'' option. ``Cannot decide'' responses are treated as missing and excluded from item-level aggregation. We collect responses from 60 English-proficient U.S. participants via Prolific ($\sim$\$18/hour). After excluding inattentive respondents based on the attention checks, each item retained approximately 51 valid ratings on average; we aggregate by the mean to obtain HGT labels rescaled to $[0,1]$. The study used anonymous online judgments on non-sensitive public image--text pairs, and no personally identifying information was collected beyond Prolific recruitment metadata. Given the small label set, HGT provides exploratory calibration evidence rather than confirmatory inference.

\noindent\textbf{Calibration Metrics.}
We assess alignment between model scores and HGT using (i) Pearson correlation ($Corr$); (ii) mean absolute error $\mathrm{MAE}=\frac{1}{N}\sum_i |CS_i-HGT_i|$ and mean squared error $\mathrm{MSE}=\frac{1}{N}\sum_i (CS_i-HGT_i)^2$, where MSE is used here as a continuous-target calibration error against HGT; (iii) a binned calibration curve with equal-frequency bins over predicted CS (we use $M{=}5$ bins for $N{=}25$); and (iv) Orthogonal Deming Regression (ODR), estimating $HGT=\beta_0+\beta_1 CS + e$. When $\hat{\beta}_0$ is not statistically different from zero, we also fit the restricted model $HGT=\beta_1 CS + e$. We report standard errors and $p$-values (*$p{<}.05$, **$p{<}.01$, ***$p{<}.001$). We do not report Expected Calibration Error (ECE): ECE compares predicted confidence against empirical binary accuracy, whereas our HGT targets are continuous. 

\subsection{Results}
\label{sec:exp_results}
Table~\ref{tab:optim_results} summarizes performance on DOCCI and Urban1k, using the convention \textit{model-dataset-k}. For $k{=}\mathrm{ZS}$, scores are zero-shot similarities from the frozen encoders; for $k{>}0$, scores come from the trained scoring layer.

\begin{figure}[t]
  \centering
  \includegraphics[width=0.95\linewidth]{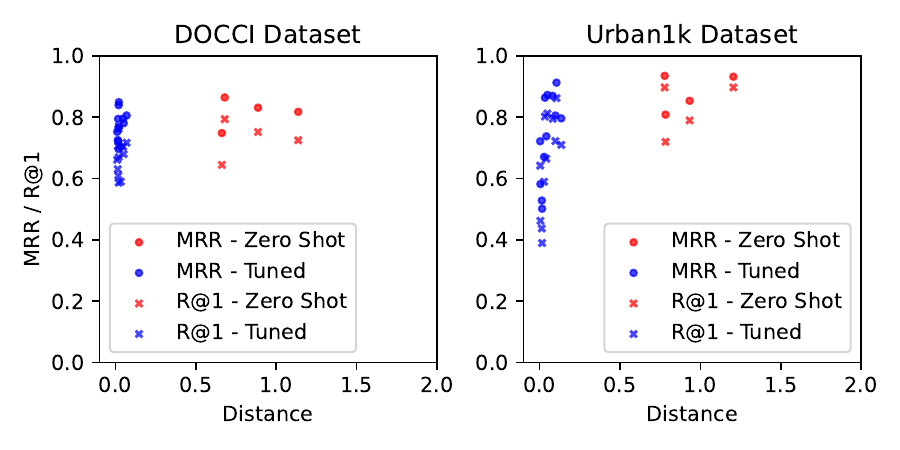}
  \caption{Distance and retrieval metrics (all backbones).}
  \label{fig:trade_off2}
\end{figure}

\noindent\textbf{Retrieval and Modality Gap.}
Figure~\ref{fig:trade_off2} relates retrieval performance to modality gap. A smaller modality gap does not uniformly imply higher retrieval performance: Jina-DOCCI-ZS outperforms Jina-DOCCI-$n^{3/4}$, while LongCLIP-DOCCI-$2n$ improves over LongCLIP-DOCCI-ZS. In several configurations, zero-shot backbones retain stronger MRR and Recall@1 than their gap-aware counterparts even when the projected embeddings exhibit a smaller centroid distance, indicating that embedding-space proximity and retrieval effectiveness capture different properties of the representation. For long-text congruence scoring this distinction matters, because a model that ranks the paired item correctly may still produce similarity scores that are poorly calibrated as absolute congruence evidence. For Urban1k, the small sample size limits the effectiveness of fitting the projection maps, so its results serve as a robustness check rather than a basis for generalization. Some non-Jina configurations show degenerate classification behavior under specific negative-sampling ratios, such as near-all-positive or near-zero-recall solutions for CLIP or SigLIP at $k=n$, and low-recall configurations on Urban1k at $k=2n$. We keep these rows in Table~\ref{tab:optim_results} for transparency, since they illustrate the instability of fitting the lightweight scoring layer outside the Jina-DOCCI setting used for the HGT calibration analysis. The statement that precision and recall remain above 95\% therefore applies only to the Jina-DOCCI configurations.

\begin{figure}[t]
  \centering
  \IfFileExists{images/final_comprehensive_calibration_plot.pdf}{%
    \includegraphics[width=0.95\linewidth]{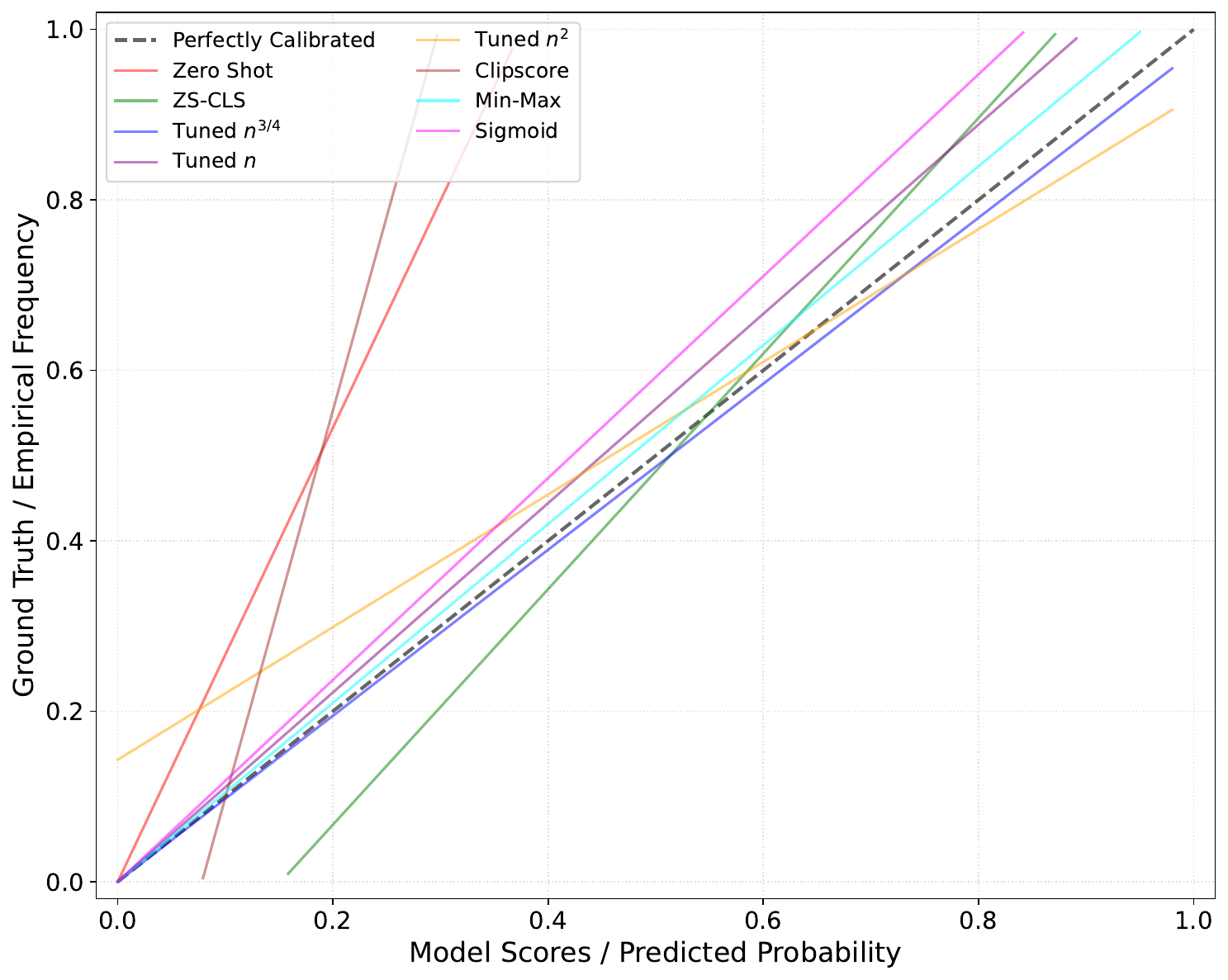}%
  }{%
    \fbox{\parbox{0.90\linewidth}{\centering
    Placeholder for calibration curves. Generate the calibration figure
    from the paired CS--HGT values before submission.}}%
  }
  \caption{Calibration curves for human-grounded congruence on DOCCI. For each model, items are sorted by predicted CS and grouped into equal-frequency bins over predicted CS scores ($M{=}5$, $N{=}25$). Each point reports the bin-wise mean CS and mean HGT, and the diagonal is the ideal $CS{=}HGT$ line. Curves are shown for CLIPScore, zero-shot similarity (ZS), ZS-CLF, and the best projection+CLF configuration.}
  \label{fig:reliability}
\end{figure}

\noindent\textbf{Human-Grounded Calibration.}
We focus on Jina-DOCCI-$k$, which achieves high precision and recall ($\geq$95\%) across sampling ratios (Table~\ref{tab:optim_results}). We organize the comparison as a calibration ladder of increasing structure: (i) CLIPScore \cite{hessel2021clipscore} and raw zero-shot similarity (ZS), uncalibrated reference-free signals; (ii) ZS-CLF, which keeps the zero-shot embeddings fixed and learns only a Platt-style affine--sigmoid calibration on the similarity logits; and (iii) the full projection+CLF models, which add the trainable projection maps before calibration. The ladder is designed to isolate the contribution of each component rather than to favor the most complex model: ZS-CLF is the critical control, because it tests whether score miscalibration can be corrected without changing the embedding geometry, and the comparison against projection+CLF therefore pinpoints whether the projection maps change threshold-relevant calibration beyond a score-only transformation. Two lightweight post-hoc normalizations of the raw ZS similarity (min--max scaling and $z$-score followed by a sigmoid) are included as additional reference points computed from existing scores. Table~\ref{tab:cs_scores} reports calibration results, and Figure~\ref{fig:reliability} shows calibration curves. The $z$-sigmoid normalization is included as a lightweight post-hoc reference, but its transformed scores yield a near-degenerate distribution for ODR; we therefore report its $Corr$, MAE, MSE, and paired test, but omit ODR estimates.
Raw signals are miscalibrated relative to human judgments. Jina-DOCCI-ZS is underconfident, with ODR slope $\hat{\beta_1}=2.66$ (SE $0.12$, $\hat{\beta_1}>1$); CLIPScore is likewise underconfident ($\hat{\beta_0}\neq 0$, $\hat{\beta_1}>1$),
all requiring calibration before their magnitude can be read as congruence evidence.

\noindent\textbf{Reconciling Retrieval, Rank Association, and Calibration.}
These results expose a trade-off rather than a uniform improvement. The ZS-CLF ablation, which calibrates directly on zero-shot similarity without the projection maps, attains the highest association with HGT ($Corr{=}0.93$) and a non-significant mean difference ($t{=}0.32$), yet retains a statistically significant intercept ($\hat{\beta}_0{=}{-}0.21$, $p{<}.001$). This indicates that post-hoc monotonic calibration already corrects mean-level misalignment but leaves a systematic offset in the score--HGT mapping. Adding the projection maps lowers association with HGT (e.g., $Corr$ falls to 0.80 and 0.65 for $k{=}n^{3/4}$ and $k{=}n$) and reduces retrieval performance relative to zero-shot retrieval (Table~\ref{tab:optim_results}). Its defensible contribution is therefore narrower: for $k{=}n^{3/4}$ and $k{=}n$, it moves the restricted ODR slope closer to 1 and removes statistically significant intercept offset. When a score is used as a thresholdable evaluation signal rather than a ranking function, systematic offset can be more consequential than rank association, because intercept shifts produce persistent over- or under-screening at fixed thresholds. This matters specifically for long-text congruence use cases that consume the score rather than the ranking: screening flags pairs below a fixed congruence cutoff, filtering removes weakly grounded image--text matches, and cross-instance comparison assumes that equal scores denote equal congruence. A high-correlation but offset-biased score, as produced by ZS-CLF, preserves the ordering needed for retrieval yet systematically mislocates the decision boundary for all three, whereas a configuration that removes the significant intercept and moves the restricted slope toward one yields a score whose absolute magnitude is more directly actionable at a threshold. The projection maps are therefore not a uniform improvement but a targeted intervention on the score property that ranking metrics and correlation do not capture.

\noindent\textbf{ZS-CLF as a Critical Control.}
The strong performance of ZS-CLF is not a failure case for the proposed framework, but a central diagnostic result. It shows that a large portion of human association and pointwise calibration error can be corrected by a score-only affine--sigmoid transformation without changing the embedding geometry, {\color{black} in our case reducing the modality gap.}
The projection+CLF variants therefore should not be read as uniformly superior replacements for ZS-CLF. Their role is narrower and diagnostic: they test whether {\color{black} closing the modality gap} alters threshold-relevant calibration beyond what a monotonic score transformation can achieve. The resulting pattern clarifies the distinction among three objectives that are often conflated in reference-free image--text evaluation: preserving retrieval rank, maximizing association with human judgments, and producing a score whose intercept and slope support fixed-threshold use. Under this interpretation, ZS-CLF is the strongest score-only calibration baseline, while projection+CLF exposes when geometry-level intervention changes threshold behavior at the cost of association and retrieval.

\noindent\textbf{Sensitivity of the Calibration Evidence.}
Because HGT contains aggregated item-level labels, we treat the calibration analysis as exploratory rather than confirmatory. The five perturbed mid-range items were included to avoid a purely binary positive--negative score distribution and to expose calibration behavior in the intermediate range. We therefore interpret the main evidence through the point estimates in Table~\ref{tab:cs_scores} and the qualitative calibration pattern in Figure~\ref{fig:reliability}. This design provides initial evidence of the retrieval--association--calibration trade-off, but larger HGT sets are required for stable confidence intervals and stronger claims about generalization.

\begin{table}[t]
  \centering
  \caption{Optimization results for DOCCI and Urban1k. We report image-to-text retrieval metrics (MRR and R@1), and positive/negative classification precision and recall for each backbone and negative-sampling setting.}
  \label{tab:optim_results}
  \scriptsize
  \setlength{\tabcolsep}{3pt}
  \begin{tabular}{lll|ll|l|ll}
    \toprule
    Dataset & Backbone & Model & MRR & R@1 & Distance & Precision & Recall \\
    \midrule
    DOCCI & Jina & ZS & 0.8601 & 0.7876 & 0.6795 & - & - \\
     & & $k=n^{3/4}$ & 0.7916 & 0.7030 & 0.0162 & 0.9953 & 0.9738 \\
     & & $k=n$ & 0.8099 & 0.7214 & 0.0690 & 0.9656 & 0.9988 \\
     & & $k=2n$ & 0.8366 & 0.7590 & 0.0195 & 0.9977 & 0.9750 \\
     & CLIP & ZS & 0.7599 & 0.6576 & 0.6616 & - & - \\
     & & $k=n^{3/4}$ & 0.7192 & 0.6058 & 0.0174 & 0.9931 & 0.9246 \\
     & & $k=n$ & 0.6985 & 0.5828 & 0.0336 & 0.5000 & 1.0000 \\
     & & $k=2n$ & 0.6992 & 0.5898 & 0.0191 & 0.9970 & 0.9172 \\
     & SigLIP & ZS & 0.8269 & 0.7446 & 0.8864 & - & - \\
     & & $k=n^{3/4}$ & 0.7216 & 0.6230 & 0.0141 & 0.9629 & 0.9816 \\
     & & $k=n$ & 0.7697 & 0.6726 & 0.0198 & 0.4995 & 0.0003 \\
     & & $k=2n$ & 0.7490 & 0.6550 & 0.0105 & 0.9989 & 0.8724 \\
     & LongCLIP & ZS & 0.7767 & 0.6654 & 1.1365 & - & - \\
     & & $k=n^{3/4}$ & 0.7918 & 0.6988 & 0.0468 & 0.9953 & 0.9702 \\
     & & $k=n$ & 0.7782 & 0.6768 & 0.0513 & 0.9985 & 0.5454 \\
     & & $k=2n$ & 0.8424 & 0.7598 & 0.0209 & 0.9979 & 0.9696 \\
    \midrule
    Urban1k & Jina & ZS & 0.9297 & 0.8800 & 0.7781 & - & - \\
     & & $k=n^{3/4}$ & 0.8639 & 0.8050 & 0.0335 & 0.9419 & 0.7300 \\
     & & $k=n$ & 0.9120 & 0.8600 & 0.1052 & 0.8905 & 0.8950 \\
     & & $k=2n$ & 0.7303 & 0.6600 & 0.0038 & 0.9726 & 0.3550 \\
     & CLIP & ZS & 0.8435 & 0.7550 & 0.7834 & - & - \\
     & & $k=n^{3/4}$ & 0.6758 & 0.5950 & 0.0279 & 0.6970 & 0.6900 \\
     & & $k=n$ & 0.7955 & 0.7100 & 0.1351 & 0.3294 & 0.9750 \\
     & & $k=2n$ & 0.5307 & 0.4450 & 0.0139 & 0.9787 & 0.2300 \\
     & SigLIP & ZS & 0.8602 & 0.7950 & 0.9333 & - & - \\
     & & $k=n^{3/4}$ & 0.7361 & 0.6550 & 0.0420 & 0.8353 & 0.7100 \\
     & & $k=n$ & 0.8206 & 0.7450 & 0.0985 & 0.6204 & 0.8500 \\
     & & $k=2n$ & 0.5087 & 0.4050 & 0.0157 & 0.9600 & 0.1200 \\
     & LongCLIP & ZS & 0.9220 & 0.8750 & 1.2056 & - & - \\
     & & $k=n^{3/4}$ & 0.8740 & 0.8050 & 0.0486 & 0.7302 & 0.9200 \\
     & & $k=n$ & 0.8638 & 0.7950 & 0.0818 & 0.8662 & 0.6800 \\
     & & $k=2n$ & 0.5826 & 0.4550 & 0.0054 & 0.9516 & 0.2950 \\
    \bottomrule
  \end{tabular}
\end{table}
\begin{table*}[t]
    \centering
    \caption{CS--HGT calibration for JinaCLIP-DOCCI. MAE and MSE are mean absolute error and mean squared error against the aggregated HGT labels; MSE is used as a continuous-target calibration error.}
    \label{tab:cs_scores}
    \small
    \setlength{\tabcolsep}{5pt}
    \begin{tabular}{lcccclc}
        \toprule
        Model & Corr & MAE & MSE & T-test & ODR $\beta_0$, $\beta_1$ & Restricted ODR $\beta_1$ \\
        \midrule
        CLIPScore \cite{hessel2021clipscore} & 0.90*** & 0.35 & 0.19 & -5.31*** & -0.35***, 4.54*** & -- \\
         & & & & & (0.09, 0.44) & -- \\
        \midrule
        ZS & 0.92*** & 0.32 & 0.16 & -6.12*** & -0.05, 2.84*** & 2.66*** \\
         & & & & & (0.06, 0.24) & (0.12) \\
        ZS + min--max & 0.92*** & 0.11 & 0.02 & -0.61 & -0.01, 1.07*** & 1.04*** \\
         & & & & & (0.05, 0.09) & (0.05) \\
        ZS + $z$-sigmoid & 0.92*** & 0.32 & 0.12 & 0.54 & -- & -- \\
        & & & & & -- & -- \\
        ZS-CLF & 0.93*** & 0.08 & 0.014 & 0.32 & -0.21***, 1.38*** & -- \\
         & & & & & (0.06, 0.11) & -- \\
        $k=n^{3/4}$ & 0.80*** & 0.18 & 0.08 & -0.79 & 0.14, 0.79*** & 0.97*** \\
         & & & & & (0.07, 0.11) & (0.08) \\
        $k=n$ & 0.65*** & 0.20 & 0.08 & -0.03 & -0.31, 1.62*** & 1.11*** \\
         & & & & & (0.18, 0.32) & (0.10) \\
        $k=2n$ & 0.82*** & 0.18 & 0.07 & -0.71 & 0.14*, 0.78*** & -- \\
         & & & & & (0.07, 0.10) & -- \\
        \bottomrule
    \end{tabular}
    \begin{flushleft}
    \footnotesize
    Notes: Restricted ODR estimates report $\beta_1$ from $HGT=\beta_1 CS+e$ only when the intercept in the unrestricted ODR model is not statistically different from zero. Parentheses report standard errors for ODR estimates. The unrestricted ODR fit for ZS+$z$-sigmoid is not reported because the transformed scores yield a numerically unstable fit. HGT contains aggregated item-level labels. *$p{<}.05$, **$p{<}.01$, ***$p{<}.001$.
    \end{flushleft}
\end{table*}

\section{Discussion and Limitations}
\label{sec:design_implications}
The central finding is that calibrated congruence scoring is a distinct evaluation objective for long-text vision-language systems. A score used for retrieval only needs to order candidates correctly; a score used for congruence screening must also have meaningful numerical scale. Our results show that these objectives can diverge: the configurations with the strongest association or retrieval behavior are not necessarily those with the most appropriate threshold geometry.

The contribution of CS is therefore best understood as making four separations explicit. First, it separates long-text congruence scoring from standard image--caption retrieval: the relevant output is a calibrated pairwise score, not only a ranking over candidates. Second, it separates post-projection representation proximity from retrieval effectiveness: our results across four frozen backbones show that observed reductions in centroid distance do not mechanically improve MRR or Recall@1. Third, it separates raw dual-encoder similarity from interpretable congruence: CLIPScore and zero-shot similarities can remain strongly associated with HGT while still exhibiting underconfidence or intercept offset. Fourth, it separates model-side optimization from human-grounded evaluation: the final score is assessed against aggregated human judgments rather than only against synthetic positive and negative labels. These separations are important because deployed long-text settings often consume scores directly. A fixed threshold for screening weak image--text matches, flagging unsupported claims, or comparing generated descriptions across cases depends on the numerical meaning of the score, not only on retrieval rank.
The backbone analysis shows that CS can be instantiated across heterogeneous vision-language encoders while preserving the distinction between representation geometry, retrieval behavior, and score calibration. The four backbones differ in context length, embedding dimension, and contrastive training family, and the experiments show that fitting a lightweight scoring layer can substantially change centroid distance and classification behavior without modifying the frozen encoders. The human-grounded analysis is intentionally concentrated on Jina-DOCCI, where the backbone supports long text and the learned scoring layer achieves stable positive/negative discrimination. This division of evidence reflects the paper's goal: broad backbone experiments establish the retrieval--geometry trade-off, while the HGT study examines whether calibrated scores behave meaningfully against human judgments.

Few limitations bound these claims. First, the human-grounded evaluation rests on aggregated labels and should be read as exploratory calibration evidence rather than confirmatory inference. 
Second, we evaluate two long-text datasets and a single linear projection layer plus Platt-style calibration; stronger calibration baselines such as held-out temperature scaling, isotonic regression, spline calibration, and beta calibration, and alternative projections (CCA, orthogonal Procrustes) are left to future work. Third, we treat the negative sampling ratio as a fixed hyperparameter. 
{\color{black} Fourth, we acknowledge the small evaluation human-grounded dataset of 25 pairs. Gathering human feedback is costly, but richer data collection on sample size and different dataset would be warranted, which we leave for future work.}

\section{Conclusion}
\label{sec:conclusions}
This paper studied long-text image--text congruence scoring under the modality gap. Across DOCCI and Urban1k, observed reductions in post-projection centroid distance did not uniformly improve retrieval performance, showing that retrieval effectiveness and embedding-space proximity are not mechanically coupled. Human-grounded calibration results further show that raw similarity and CLIPScore can be underconfident relative to human judgments, while projection-based configurations can reduce slope and intercept distortions in selected settings. The central finding is therefore a trade-off: CS does not uniformly improve retrieval performance or association with HGT, but exposes a tension between conventional retrieval and correlation metrics and threshold calibration. Future work should expand human-grounded labels, test additional calibration baselines, and evaluate the method across broader vision-language tasks such as captioning and visual question answering.

\printbibliography

@inproceedings{krishna2017visualgenome,
  title     = {Visual Genome: Connecting Language and Vision Using Crowdsourced Dense Image Annotations},
  author    = {Krishna, Ranjay and Zhu, Yuke and Groth, Oliver and Johnson, Justin and Hata, Kenji and Kravitz, Joshua and Chen, Stephanie and Kalantidis, Yannis and Li, Li-Jia and Shamma, David A. and Bernstein, Michael S. and Fei-Fei, Li},
  journal   = {International Journal of Computer Vision},
  volume    = {123},
  number    = {1},
  pages     = {32--73},
  year      = {2017}
}

@inproceedings{plummer2015flickr30kentities,
  title     = {Flickr30k Entities: Collecting Region-to-Phrase Correspondences for Richer Image-to-Sentence Models},
  author    = {Plummer, Bryan A. and Wang, Liwei and Cervantes, Chris M. and Caicedo, Juan C. and Hockenmaier, Julia and Lazebnik, Svetlana},
  booktitle = {Proceedings of the IEEE International Conference on Computer Vision (ICCV)},
  pages     = {2641--2649},
  year      = {2015}
}

@inproceedings{ponttuset2020localized,
  title     = {Connecting Vision and Language with Localized Narratives},
  author    = {Pont-Tuset, Jordi and Uijlings, Jasper and Changpinyo, Soravit and Soricut, Radu and Ferrari, Vittorio},
  booktitle = {European Conference on Computer Vision (ECCV)},
  pages     = {647--664},
  year      = {2020}
}

@inproceedings{thrush2022winoground,
  title     = {Winoground: Probing Vision and Language Models for Visio-Linguistic Compositionality},
  author    = {Thrush, Tristan and Jiang, Ryan and Bartolo, Max and Singh, Amanpreet and Williams, Adina and Kiela, Douwe and Ross, Candace},
  booktitle = {Proceedings of the IEEE/CVF Conference on Computer Vision and Pattern Recognition (CVPR)},
  pages     = {5238--5248},
  year      = {2022}
}

@inproceedings{yuksekgonul2022aro,
  title     = {When and Why Vision-Language Models Behave Like Bags-of-Words, and What to Do About It?},
  author    = {Yuksekgonul, Mert and Bianchi, Federico and Kalluri, Pratyusha and Jurafsky, Dan and Zou, James},
  booktitle = {International Conference on Learning Representations (ICLR)},
  year      = {2023}
}

@inproceedings{papineni2002bleu,
  title     = {BLEU: A Method for Automatic Evaluation of Machine Translation},
  author    = {Papineni, Kishore and Roukos, Salim and Ward, Todd and Zhu, Wei-Jing},
  booktitle = {Proceedings of the 40th Annual Meeting of the Association for Computational Linguistics (ACL)},
  pages     = {311--318},
  year      = {2002}
}

@inproceedings{vedantam2015cider,
  title     = {CIDEr: Consensus-Based Image Description Evaluation},
  author    = {Vedantam, Ramakrishna and Lawrence Zitnick, C. and Parikh, Devi},
  booktitle = {Proceedings of the IEEE Conference on Computer Vision and Pattern Recognition (CVPR)},
  pages     = {4566--4575},
  year      = {2015}
}

@inproceedings{anderson2016spice,
  title     = {SPICE: Semantic Propositional Image Caption Evaluation},
  author    = {Anderson, Peter and Fernando, Basura and Johnson, Mark and Gould, Stephen},
  booktitle = {European Conference on Computer Vision (ECCV)},
  pages     = {382--398},
  year      = {2016}
}

@inproceedings{jiang2019tiger,
  title     = {TIGEr: Text-to-Image Grounding for Image Caption Evaluation},
  author    = {Jiang, Ming and Huang, Qiuyuan and Zhang, Lei and Wang, Xin and Zhang, Pengchuan and Gan, Zhe and Diesner, Jana and Gao, Jianfeng},
  booktitle = {Proceedings of the 2019 Conference on Empirical Methods in Natural Language Processing (EMNLP)},
  pages     = {2141--2152},
  year      = {2019}
}

@inproceedings{zadrozny2002transforming,
  title     = {Transforming Classifier Scores into Accurate Multiclass Probability Estimates},
  author    = {Zadrozny, Bianca and Elkan, Charles},
  booktitle = {Proceedings of the Eighth ACM SIGKDD International Conference on Knowledge Discovery and Data Mining (KDD)},
  pages     = {694--699},
  year      = {2002}
}

@inproceedings{niculescu2005predicting,
  title     = {Predicting Good Probabilities with Supervised Learning},
  author    = {Niculescu-Mizil, Alexandru and Caruana, Rich},
  booktitle = {Proceedings of the 22nd International Conference on Machine Learning (ICML)},
  pages     = {625--632},
  year      = {2005}
}

@inproceedings{naeini2015obtaining,
  title     = {Obtaining Well Calibrated Probabilities Using Bayesian Binning},
  author    = {Naeini, Mahdi Pakdaman and Cooper, Gregory F. and Hauskrecht, Milos},
  booktitle = {Proceedings of the Twenty-Ninth AAAI Conference on Artificial Intelligence},
  pages     = {2901--2907},
  year      = {2015}
}

@inproceedings{guo2017calibration,
  title     = {On Calibration of Modern Neural Networks},
  author    = {Guo, Chuan and Pleiss, Geoff and Sun, Yu and Weinberger, Kilian Q.},
  booktitle = {Proceedings of the 34th International Conference on Machine Learning (ICML)},
  pages     = {1321--1330},
  year      = {2017}
}

@inproceedings{kuleshov2018calibrated,
  title     = {Accurate Uncertainties for Deep Learning Using Calibrated Regression},
  author    = {Kuleshov, Volodymyr and Fenner, Nathan and Ermon, Stefano},
  booktitle = {Proceedings of the 35th International Conference on Machine Learning (ICML)},
  pages     = {2796--2804},
  year      = {2018}
}

@inproceedings{hessel2021clipscore,
  title={CLIPScore: A Reference-free Evaluation Metric for Image Captioning},
  author={Hessel, Jack and Holtzman, Ari and Forbes, Maxwell and Le Bras, Ronan and Choi, Yejin},
  booktitle={Proceedings of the 2021 Conference on Empirical Methods in Natural Language Processing},
  pages={7514--7528},
  year={2021}
}

@inproceedings{steck2024cosine,
  title={Is cosine-similarity of embeddings really about similarity?},
  author={Steck, Harald and Ekanadham, Chaitanya and Kallus, Nathan},
  booktitle={Companion Proceedings of the ACM on Web Conference 2024},
  pages={887--890},
  year={2024}
}

@inproceedings{shi2023towards,
  title={Towards understanding the modality gap in CLIP},
  author={Shi, Peiyang and Welle, Michael C and Bj{\"o}rkman, M{\aa}rten and Kragic, Danica},
  booktitle={ICLR 2023 Workshop on Multimodal Representation Learning: Perks and Pitfalls},
  year={2023}
}

@inproceedings{jiang2023understanding,
  title={Understanding and constructing latent modality structures in multi-modal representation learning},
  author={Jiang, Qian and Chen, Changyou and Zhao, Han and Chen, Liqun and Ping, Qing and Tran, Son Dinh and Xu, Yi and Zeng, Belinda and Chilimbi, Trishul},
  booktitle={Proceedings of the IEEE/CVF Conference on Computer Vision and Pattern Recognition},
  pages={7661--7671},
  year={2023}
}

@article{li2021align,
  title={Align before fuse: Vision and language representation learning with momentum distillation},
  author={Li, Junnan and Selvaraju, Ramprasaath and Gotmare, Akhilesh and Joty, Shafiq and Xiong, Caiming and Hoi, Steven Chu Hong},
  journal={Advances in Neural Information Processing Systems},
  volume={34},
  pages={9694--9705},
  year={2021}
}

@inproceedings{li2022blip,
  title={Blip: Bootstrapping language-image pre-training for unified vision-language understanding and generation},
  author={Li, Junnan and Li, Dongxu and Xiong, Caiming and Hoi, Steven},
  booktitle={International Conference on Machine Learning},
  pages={12888--12900},
  year={2022},
  organization={PMLR}
}

@article{koukounas2024jina,
  title={Jina CLIP: Your CLIP Model Is Also Your Text Retriever},
  author={Koukounas, Andreas and Mastrapas, Georgios and G{\"u}nther, Michael and Wang, Bo and Martens, Scott and Mohr, Isabelle and Sturua, Saba and Akram, Mohammad Kalim and Mart{\'\i}nez, Joan Fontanals and Ognawala, Saahil and others},
  journal={arXiv preprint arXiv:2405.20204},
  year={2024}
}

@inproceedings{radford2021learning,
  title={Learning transferable visual models from natural language supervision},
  author={Radford, Alec and Kim, Jong Wook and Hallacy, Chris and Ramesh, Aditya and Goh, Gabriel and Agarwal, Sandhini and Sastry, Girish and Askell, Amanda and Mishkin, Pamela and Clark, Jack and others},
  booktitle={International Conference on Machine Learning},
  pages={8748--8763},
  year={2021},
  organization={PMLR}
}

@inproceedings{zhai2023sigmoid,
  title={Sigmoid loss for language image pre-training},
  author={Zhai, Xiaohua and Mustafa, Basil and Kolesnikov, Alexander and Beyer, Lucas},
  booktitle={Proceedings of the IEEE/CVF International Conference on Computer Vision},
  pages={11975--11986},
  year={2023}
}

@inproceedings{zhang2024long,
  title={Long-clip: Unlocking the long-text capability of clip},
  author={Zhang, Beichen and Zhang, Pan and Dong, Xiaoyi and Zang, Yuhang and Wang, Jiaqi},
  booktitle={European Conference on Computer Vision},
  pages={310--325},
  year={2024},
  organization={Springer}
}

@inproceedings{onoe2024docci,
  author        = {Yasumasa Onoe and Sunayana Rane and Zachary Berger and Yonatan Bitton and Jaemin Cho and Roopal Garg and
    Alexander Ku and Zarana Parekh and Jordi Pont-Tuset and Garrett Tanzer and Su Wang and Jason Baldridge},
  title         = {{DOCCI: Descriptions of Connected and Contrasting Images}},
  booktitle     = {ECCV},
  year          = {2024}
}

@article{liang2022mind,
  title={Mind the gap: Understanding the modality gap in multi-modal contrastive representation learning},
  author={Liang, Victor Weixin and Zhang, Yuhui and Kwon, Yongchan and Yeung, Serena and Zou, James Y},
  journal={Advances in Neural Information Processing Systems},
  volume={35},
  pages={17612--17625},
  year={2022}
}

@article{mikolov2013distributed,
  title={Distributed representations of words and phrases and their compositionality},
  author={Mikolov, Tomas and Sutskever, Ilya and Chen, Kai and Corrado, Greg S and Dean, Jeff},
  journal={Advances in Neural Information Processing Systems},
  volume={26},
  year={2013}
}

@inproceedings{robinson2021contrastive,
  title={Contrastive Learning With Hard Negative Samples},
  author={Robinson, Joshua and Chuang, Ching-Yao and Sra, Suvrit and Jegelka, Stefanie},
  booktitle={International Conference on Learning Representations (ICLR)},
  year={2021}
}

@article{wang2025cross,
  title={Cross-modal retrieval: a systematic review of methods and future directions},
  author={Wang, Tianshi and Li, Fengling and Zhu, Lei and Li, Jingjing and Zhang, Zheng and Shen, Heng Tao},
  journal={Proceedings of the IEEE},
  year={2025},
  publisher={IEEE}
}

@incollection{platt1999probabilistic,
  title     = {Probabilistic outputs for support vector machines and comparisons to regularized likelihood methods},
  author    = {Platt, John},
  booktitle = {Advances in Large Margin Classifiers},
  editor    = {Smola, Alexander J. and Bartlett, Peter and Sch{\"o}lkopf, Bernhard and Schuurmans, Dale},
  volume    = {10},
  number    = {3},
  pages     = {61--74},
  year      = {1999},
  publisher = {MIT Press}
}

@inproceedings{urbanek2024picture,
  title={A picture is worth more than 77 text tokens: Evaluating clip-style models on dense captions},
  author={Urbanek, Jack and Bordes, Florian and Astolfi, Pietro and Williamson, Mary and Sharma, Vasu and Romero-Soriano, Adriana},
  booktitle={Proceedings of the IEEE/CVF Conference on Computer Vision and Pattern Recognition},
  pages={26700--26709},
  year={2024}
}

@article{fahim2024modgap,
  title={It's Not a Modality Gap: Characterizing and Addressing the Contrastive Gap},
  author={Fahim, Abrar and Murphy, Alex and Fyshe, Alona},
  journal={arXiv preprint arXiv:2405.18570},
  year={2024}
}
\end{document}